\pdfoutput=1

\documentclass[11pt]{article}

\usepackage[preprint]{acl}

\usepackage{times}
\usepackage{latexsym}
 \usepackage{microtype}
\usepackage{algorithm}
\usepackage{algorithmic}
\usepackage{amsmath} 
\usepackage{tcolorbox}
\usepackage{amsfonts}
\usepackage{booktabs}      
\usepackage{multirow}      
\usepackage{caption}       
\usepackage{array}         
\usepackage[utf8]{inputenc}

\usepackage{microtype}

\usepackage{inconsolata}

\usepackage{graphicx}
\usepackage{pgfplots, pgfplotstable}
\pgfplotsset{compat=1.18}
\usepackage{booktabs}

\title{Simple Yet Effective: An Information-Theoretic Approach to Multi-LLM Uncertainty Quantification}

\author{
  \textbf{Maya Kruse\textsuperscript{1}},
  \textbf{Majid Afshar\textsuperscript{3}},
  \textbf{Saksham Khatwani\textsuperscript{1,2}},
  \textbf{Anoop Mayampurath\textsuperscript{3}},
  \textbf{Guanhua Chen\textsuperscript{3}}
\\
  \textbf{Yanjun Gao\textsuperscript{1}\thanks{Correspondence: \href{mailto:yanjun.gao@cuanschutz.edu}{yanjun.gao@cuanschutz.edu}}}
\\
\textsuperscript{1}University of Colorado Anschutz Medical Campus \\
  \textsuperscript{2}University of Colorado Boulder \\
  \textsuperscript{3}University of Wisconsin Madison\\
\\
  }

\begin{document}

\maketitle
\begin{abstract}
Large language models (LLMs) often behave inconsistently across inputs, indicating uncertainty and motivating the need for its quantification in high-stakes settings. Prior work on calibration and uncertainty quantification often focuses on individual models, overlooking the potential of model diversity. We hypothesize that LLMs make complementary predictions due to differences in training and the Zipfian nature of language, and that aggregating their outputs leads to more reliable uncertainty estimates. To leverage this, we propose \textsc{MUSE} (Multi-LLM Uncertainty via Subset Ensembles), a simple information-theoretic method that uses Jensen-Shannon Divergence to identify and aggregate well-calibrated subsets of LLMs. Experiments on binary prediction tasks demonstrate improved calibration and predictive performance compared to single-model and naïve ensemble baselines. In addition, we explore using MUSE as guided signals with chain-of-thought distillation to fine-tune LLMs for calibration. \textsc{MUSE} is available at:\url{https://github.com/LARK-NLP-Lab/MUSE}.
\end{abstract}

\section{Introduction}
\label{sec:intro}
Although large language models (LLMs) have shown remarkable performance in a wide range of NLP tasks and domains, their output is not always consistent or reliable~\cite{xiao2022uncertainty,zhao2024improving}. The same LLM can generate divergent responses under different decoding settings, even with identical inputs~\cite{wang2024integrate,wei2022chain}. As LLMs enter high-stakes domains like healthcare, quantifying output variance is essential for trust, safety, and decision-making~\cite{gaoposition,savage2025large,qin2024enhancing}.

Quantifying uncertainty is essential to address this challenge: Generating responses with appropriately calibrated confidence helps determine when the answer is trustworthy~\cite{geng2024survey}. Although prior work has explored uncertainty estimation and calibration through sampling and self-consistency~\cite{rivera2024combining,gao2024spuq,ling2024uncertainty}, uncertainty-aware training~\cite{liu2024can,chen2024quantifying,kapoor2024calibration}, reflection~\cite{zhao2024fact,zhang-etal-2024-self-contrast}, ranking~\cite{huang2024uncertainty}, and conformal prediction~\cite{wang2024conu}, these methods focus on single LLMs. 

This paper introduces a novel approach to uncertainty quantification by aggregating predictions from multiple LLMs. Different LLMs generalize better in distinct regions of the input space, due to the Zipfian nature of language and differences in training corpora, objectives, and architectures~\cite{piantadosi2014zipf,chan2022data}. Based on this, we \textit{hypothesize that} combining their outputs offers a principled way to reduce uncertainty, improve robustness, and better approximate ground truth in regions where individual models may falter. 

Specifically, we formulate the problem through an information-theoretic lens, using Jensen-Shannon Divergence (JSD) to capture the degree of disagreement among models. JSD offers a symmetric and bounded measure of divergence between probability distributions~\cite{cover1999elements}, making it well-suited for comparing predictions across multiple LLMs. We quantify model disagreement to identify reliable consensus and propose \textsc{Muse} (\textbf{M}ulti-LLM \textbf{U}ncertainty via \textbf{S}ubset \textbf{E}nsembles), a simple algorithm that selects and aggregates LLM outputs to balance diversity and reliability.

We evaluate \textsc{Muse} on three publicly available binary prediction datasets. \textit{TruthfulQA} (TQA) covers general domain questions and answers designed to investigate the truthfulness of the model~\cite{lin2022truthfulqa}; both \textit{EHRShot}~\cite{wornow2023ehrshot}, and \textit{MIMIC-Extract} (MIMIC) are structured clinical datasets derived from records of hospitalized patients in the real world~\cite{johnson2016mimic, wang2020mimic}. Focusing on binary prediction enables straightforward evaluation of both discrimination and calibration, and empirical findings provide evidence to support our hypothesis with \textsc{Muse} improving calibration and robustness.

We also ask whether consensus-driven probabilities can \textit{teach probabilistic reasoning} to individual LLMs. Using \textsc{Muse} outputs as silver-standard supervision for fine-tuning and CoT distillation, we find that ensemble-derived signals are principled but their effectiveness depends on the underlying model, highlighting an important direction for understanding how LLMs internalize probabilistic reasoning.  

\section{Related Work}

In addition to the related work discussed in \S\ref{sec:intro}, \citet{ling2024uncertainty} estimate both aleatoric and epistemic uncertainty using entropy within single-LLM in-context learning. 
\citet{chen-etal-2025-uncertainty} focus on clinical prediction tasks, applying deep ensembles and Monte Carlo dropout to capture uncertainty from a single decoder. While these methods operate within a single-model setting, our work addresses uncertainty in a multi-LLM context. In this space, \citet{zhang2024luq} quantify uncertainty across LLMs via semantic similarity in long-form generation, and \citet{dey2025uncertainty} select LLMs from a pool to reduce hallucinations based on task accuracy. In contrast, we propose an information-theoretic framework that selects LLM subsets by minimizing predictive uncertainty via JSD and entropy.  

\section{Methods}
\subsection{LLM Uncertainty Quantification}
\label{sec:single_llm}
We establish two methods for uncertainty quantification for a single LLM as: 
(1) \textit{self-consistency-based empirical estimation}, which forms the core of our proposed methods, and 
(2) \textit{sequence likelihood scoring}, used as a widely adopted baseline in prior work~\cite{geng2024survey}.

\noindent \textbf{(1) Self-Consistency with Empirical Frequency.}
Given a binary classification input, we perform stochastic decoding runs $k$ in LLM text generation (GEN), using temperature $T$ sampling ($T = 0.7$ and $k=10$), resulting in a set of outputs $\{ \hat{y_i} \}_{i=1}^k$. Each output is mapped to a binary label (\texttt{yes} or \texttt{no}). Define the empirical probability of the label \texttt{yes} as: $\hat{p}^{\texttt{yes}} = \frac{1}{k} \sum_{i=1}^k \mathbb{I}(\hat{y_i} = \texttt{yes}), 
\quad 
\hat{p}^{\texttt{no}} = 1 - \hat{p}^{\texttt{yes}}$.


To estimate uncertainty, we apply a bootstrapping procedure: we resample $90\%$ of the outputs with replacement and recompute $\hat{p}^{\texttt{yes}}$ over $B=100$ trials (denoting as GEN$^{BS}$). From the resulting $\hat{p}^{\texttt{yes}}$ distribution, we compute variance, entropy, and JSD for our proposed algorithms (\S~\ref{sec:multi_llm}). 

\noindent \textbf{(2) Sequence Likelihood (SLL) Scoring. }
We adopt the SLL approach used in prior LLM calibration work. For each input $x$, we compute the total log-likelihood of two candidate completions: \texttt{"Answer is Yes"} and \texttt{"Answer is No"}, denoted $\text{LL}_{\texttt{yes}}$ and $\text{LL}_{\texttt{no}}$. These are computed using left-to-right autoregressive decoding: $\text{LL}_{\texttt{label}} = \sum_{t=1}^{T} \log P(y_t^{\texttt{label}} \mid x, y_{<t}^{\texttt{label}})$. The final prediction probability is obtained via softmax normalization. We use this predicted distribution to compute AUROC, ECE and Brier Scores~\cite{guo2017calibration}. Although not robust to LLM output variability, SLL provides a deterministic scoring baseline for comparison.

\subsection{Multi-LLM selective algorithm}
\label{sec:multi_llm}
The central idea of \textsc{Muse} is that \textit{disagreement among LLM predictive distributions signals \textit{epistemic uncertainty}, while consensus indicates more reliable generalization.}
This can be measured by JSD~\cite{cover1999elements}. Meanwhile, we compute the mean entropy $H$ of individual model predictions to reflect \textit{aleatoric uncertainty}, capturing inherent input ambiguity. By a novel algorithm that identifies subsets of models $\mathcal{S}$ whose predictions exhibit low disagreement and low intrinsic uncertainty, this formulation enables us to surface high-consensus regions of the input space while balancing the tradeoff between diversity (which may include useful signals) and noise (which can degrade calibration and accuracy). 

\vspace{.05em}

\begin{algorithm}[t]
\footnotesize
\caption{\small \textsc{Muse}-Greedy version}
\label{alg:jsd}
\begin{algorithmic}[1]
\REQUIRE Prediction set $\mathcal{P} = \{p_i\}_{i=1}^N$, confidence $c_i = |p_i^{\text{yes}} - 0.5|$, parameters $\beta$, $\epsilon_{\text{tol}}$, $m_{\min}$
\STATE Sort $\mathcal{P}$ by $c_i$ descending; initialize $\mathcal{S} \gets \{p_1\}$, $u_{\text{epis}}^{\text{prev}} \gets 0$
\FOR{each $p_j$ in sorted $\mathcal{P} \setminus \mathcal{S}$}
  \STATE $\mathcal{S}' \gets \mathcal{S} \cup \{p_j\}$, $\bar{p} \gets \text{mean}(\mathcal{S}')$
  \STATE $u_{\text{epis}} \gets \frac{1}{|\mathcal{S}'|} \sum_{p \in \mathcal{S}'} \text{JS}(p \parallel \bar{p})^2$
  \STATE $u_{\text{alea}} \gets \frac{1}{|\mathcal{S}'|} \sum_{p \in \mathcal{S}'} H(p)$
  \IF{$|\mathcal{S}'| \geq m_{\min}$ and $u_{\text{epis}} - u_{\text{epis}}^{\text{prev}} > \epsilon_{\text{tol}}$}
    \STATE \textbf{break}
  \ENDIF
  \STATE $\mathcal{S} \gets \mathcal{S}'$, $u_{\text{epis}}^{\text{prev}} \gets u_{\text{epis}}$
\ENDFOR
\STATE $\hat{p}^{\text{yes}} \gets \text{mean}_{p \in \mathcal{S}}(p^{\text{yes}})$, $u_{\text{total}} \gets u_{\text{epis}}^{\text{prev}} + \beta \cdot u_{\text{alea}}$
\RETURN $(\hat{p}^{\text{yes}}, u_{\text{total}}, \mathcal{S})$
\end{algorithmic}
\end{algorithm}

\noindent \textbf{Problem Setup.} 
Given an input $x$, let $\mathcal{P}_x = \{\mathbf{p}_i = (p_i, 1 - p_i)\}_{i=1}^N$ be $N$ predictive distributions from multiple LLMs and/or decoding runs, where $p_i$ denotes the predicted probability of the label \texttt{yes}. Our goal is to select a subset $\mathcal{S}_x \subseteq \mathcal{P}_x$ that yields a well-calibrated, aggregated prediction $\hat{p}$. 
\vspace{.05em}

\noindent  \textbf{Uncertainty Computation.}  
The two types of uncertainty plays an important role in the proposed algorithm. Epistemic uncertainty $\mathcal{U}_{\text{epis}}(\mathcal{S})$ 
reflects inter-model disagreement and is quantified as the average JSD  between each prediction $\mathbf{p}_i$ and the subset mean $\bar{\mathbf{p}}$:

\vspace{-.08in}
{\small 
\[
\mathcal{U}_{\text{epis}}(\mathcal{S}) = \frac{1}{|\mathcal{S}|} \sum_{i \in \mathcal{S}} \text{JS}(\mathbf{p}_i \| \bar{\mathbf{p}})
\] 
}
\vspace{-.12in}

Aleatoric uncertainty $\mathcal{U}_{\text{alea}}(\mathcal{S})$ reflects intrinsic noise and is estimated by the average binary entropy:

\vspace{-.20in}
{\small
\[
\mathcal{U}_{\text{alea}}(\mathcal{S}) = \frac{1}{|\mathcal{S}|} \sum_{i \in \mathcal{S}} H(p_i),
\] 
}
\vspace{-.10in}

where $H(p) = -p \log p - (1 - p) \log (1 - p)$.

 We focus on \textit{optimizing epistemic uncertainty}, as aleatoric uncertainty stems from inherent data noise and is not reducible via model selection. The total uncertainty of a subset of LLMs, denoted as $\mathcal{S}$, is defined as the sum of its epistemic and aleatoric components: $U(\mathcal{S}) = \mathcal{U}_{\text{epis}}(\mathcal{S}) + \beta \cdot \mathcal{U}_{\text{alea}}(\mathcal{S})$, where $\beta$ is a weighting factor that controls the trade-off between epistemic disagreement and inherent input ambiguity. Results using total uncertainty $U(\mathcal{S})$ are reported in Appendix~\ref{app:total_uncertainty}. 

\noindent \textbf{Multi-LLM Uncertainty via Subset Ensemble}.
The key contribution of this paper is \textsc{MUSE}, an algorithm that constructs well-calibrated ensembles of LLMs output based on $\mathcal{U}_{\text{epis}}(\mathcal{S})$. It supports two subset selection strategies, \textit{greedy} and \textit{conservative}, which incrementally select a subset of LLMs whose outputs are mutually diverse yet coherent, as determined by pairwise JSD. Two key parameters control the behavior of \textsc{Muse}: the noise threshold ($\epsilon_{\text{tol}}$) and the minimum subset size $m_{min}$ as diversity constraint, controlling the balance between ensemble breadth and agreement. The \textbf{greedy} version starts with the most confident LLM prediction and iteratively adds models that increase the overall $\mathcal{U}_{\text{epis}}(\mathcal{S})$ (diversity) of the subset, up to a specified tolerance (as in Algo~\ref{alg:jsd}.). The \textbf{conservative} version, in contrast, selects models that minimize a joint objective combining epistemic and aleatoric uncertainty. This approach encourages diversity while avoiding instability, resulting in a more calibrated and robust ensemble (see Algo.\ref{alg:muse_conservative}). 

Once a subset is selected, we compute the final predicted probability by averaging the individual LLM predictions within the subset. Two aggregation strategies are deployed: (1) a simple unweighted mean, and (2) an \textit{aleatoric-aware weighting}, where each LLM's prediction is $\hat{p}^{yes}$ weighted by its $\mathcal{U}_{\text{alea}}(\mathcal{S})$, where each $\hat{p}^{yes}_i$ is weighted by its entropy, i.e., $1-H(\hat{p}^{yes}_i)$. The final prediction is computed as a weighted average, assigning higher weights to more confident (low-entropy) predictions, emphasizing more decisive predictions, particularly when individual models exhibit varying uncertainty levels.

\begin{table}[t]
\centering
\small
\renewcommand{\arraystretch}{0.8}
\resizebox{\columnwidth}{!}{
\begin{tabular}{lllccc}
\toprule
\textbf{} & \textbf{LLM} & \textbf{Method} & \textbf{AUROC} & \textbf{ECE} & \textbf{Brier} \\
\midrule
\multirow{8}{*}{\textsc{single}} & \multirow{3}{*}{Mistral-7B}
  & SLL        & 64.99 & 58.11 & 55.57 \\
&  & GEN\textsuperscript{BS}     & 29.48 & 52.25 & 54.92 \\
& \multirow{3}{*}{Qwen2-7B}
  & SLL       & 58.47 & 65.54 & 65.23 \\
&  & GEN\textsuperscript{BS}    & 60.78 & 47.71 & 49.50 \\
&\multirow{3}{*}{Gemma-7B}
  & SLL       & 60.65 & 37.41 & 36.78 \\
&  & GEN\textsuperscript{BS}    & 52.14 & 48.40 & 55.88 \\
& \multirow{3}{*}{DS-Qwen-32B}
  & SLL        & \textbf{72.89} & 57.30 & 54.16 \\
&  & GEN\textsuperscript{BS}    & 63.11 & \textbf{18.83} & \textbf{30.32} \\
\midrule
\textsc{Muse} & \multirow{2}{*}{All LLMs}
& \scriptsize{mean}  & 57.57 & 38.59 & 40.68 \\
Greedy &  & \scriptsize{weigthed}   & 59.28 & 41.53 & 43.39 \\
& \multirow{2}{*}{Excl. Outlier}
 & \scriptsize{mean}  & 69.54 & 40.29 & 40.45 \\
&  & \scriptsize{weigthed}   & 68.93 & 41.11 & 41.60 \\
& DS+Qwen2  & \scriptsize{mean}   & 69.98 & 40.27 & 41.30 \\
& (Top 2)       & \scriptsize{weighted}   & 69.86 & 41.23 & 42.21 \\
\midrule
\textsc{Muse} & \multirow{2}{*}{All LLMs}
  & \scriptsize{mean} & 51.04 & 40.06 & 42.10 \\
Conserv.&  & \scriptsize{weighted} & 54.45 & 43.51 & 45.62 \\
& \multirow{2}{*}{Excl. Outlier}
  & \scriptsize{mean} & 67.57
 &39.01 & 38.48 \\
&   & \scriptsize{weighted}  & 67.30 & 40.49 & 39.99 \\
&  DS+Qwen2      & \scriptsize{mean}   & $\dagger$72.33 & $\dagger$38.15 & $\dagger$38.55 \\
 &  (Top 2)     & \scriptsize{weighted}   & 72.35 & 39.45 & 40.25 \\
\bottomrule
\end{tabular}
}
\vspace{-.1in}
\caption{\small Performance on TruthfulQA. We report SLL and GEN\textsuperscript{BS} results alongside all \textsc{Muse} settings (greedy/conservative, with or without aleatoric weighting). $\dagger$ highlight cases where \textsc{Muse} yields competitive AUROC with lower calibration error despite not being the top performer.}
\label{tab:truthfulqa_llm_results}
\end{table} 

\begin{table}[t!]
\centering
\small
\renewcommand{\arraystretch}{0.78} 
\begin{tabular}{lccc}
\toprule
\textbf{LLMs} & \textbf{AUROC} $\uparrow$ & \textbf{ECE} $\downarrow$ & \textbf{Brier Score} $\downarrow$ \\
\midrule
Qwen2 & 53.52 & 22.22 & 32.60 \\
Mistral & 60.10 & 13.70 & 25.50 \\
Gemma & 50.20 & \textbf{9.80} & 32.30 \\
DS-Qwen & 62.00 & 10.50 & 30.00 \\
\midrule
DS+Qwen2 & 61.78 & 17.50 & 28.89 \\
DS+Mistr & \textbf{64.73} & 16.57 & 25.00 \\
DS+Mistr+Qwen2 & 62.24 & 11.74 & \textbf{24.38} \\
All & 60.94 & 12.76 & 24.38 \\
\bottomrule
\end{tabular}
\vspace{-.1in}
\caption{\small Performance on the EHRShot Acute myocardial infarction (\textit{acute\_mi}) task across single LLMs (GEN\textsuperscript{BS})and multi-LLM combinations. Greedy and conservative (non-weighted) results are identical. DS = DS-Qwen.}
\label{tab:multi_llm_results}
\end{table}

\begin{table*}[t!]
\centering
\small
\renewcommand{\arraystretch}{0.6}  
\begin{tabular}{llcccccccccc}
\toprule
\textbf{LLM} & \textbf{Method} & 
\multicolumn{3}{c}{\textbf{LOS3}} & 
\multicolumn{3}{c}{\textbf{LOS7}} & 
\multicolumn{3}{c}{\textbf{Mort Hosp.}} \\
\cmidrule(r){3-5} \cmidrule(r){6-8} \cmidrule(r){9-11}
& & \textbf{AUROC} & \textbf{ECE} & \textbf{Brier} 
  & \textbf{AUROC} & \textbf{ECE} & \textbf{Brier} 
  & \textbf{AUROC} & \textbf{ECE} & \textbf{Brier} \\
\midrule
DS-Qwen & SLL               & 41.30 & 37.70 & 41.05    & 40.60 & 3.80  & 6.81    & 46.90 & 5.20  & 14.77 \\
 & GEN\textsuperscript{BS} & 57.80 & 13.90 & 31.93    & 59.10 & 30.23  &  26.82  & 55.52 & 3.57  & 11.76 \\
Qwen2-7B        & SLL          & 56.40 & 36.50 & 35.96    & 58.45 &  55.66 & 69.60 & 68.01 & 47.70 & 47.38 \\
        & GEN\textsuperscript{BS} & 56.80 & 31.10 & 34.18    & 58.45 & 45.80 & 55.66    & 59.29 & 3.30  & 10.17 \\
\midrule
Naive & Majo.    & 54.04 & \textbf{2.41}  & 38.64 & 57.42 & \textbf{7.15} & \textbf{9.18}  & 58.73 & 5.76  & 11.58 \\
 & Mean    & 53.87 & 6.98  & 40.99 & 58.40 & 3.34  & 7.76  & 59.48 & 2.28  & 10.15 \\
\midrule
\textsc{Muse} & Greedy         & \textbf{61.47} & 12.43 & \textbf{27.51}& 60.47 & 35.01 & 26.04 & 59.55 & 2.38  & 10.46 \\
& \scriptsize{+Weighted} & 61.40 & 18.83 & 29.71 & 61.29 & 34.54 & 28.22 & 59.64 & 2.29  & 10.44 \\
 & Conserv.   & 60.06 & 21.43 & 30.97 & 61.58 & 35.13 & 29.86 & 59.84 & 2.61  & 10.49 \\
& \scriptsize{+Weighted} & 61.04 & 24.03 & 32.49 & \textbf{61.71} & 35.42 & 31.39 & 59.83 & 2.76 & 10.44 \\
\bottomrule
\end{tabular}
\vspace{-.1in}
\caption{\small AUROC, ECE, and Brier Score across clinical prediction tasks. We include DS-Qwen and Qwen-7B because they are the two best-performing single LLMs on this dataset (see more results in Table~\ref{tab:mist_gem_mimic}). We compare individual LLMs, simple fusion baselines (majority voting ``Majo." and mean), and algorithmic subset selection strategies (greedy and conservative). }
\label{tab:llm_jsd_summary}
\end{table*}

\subsection{Uncertainty-Aware Supervised Fine-Tuning}
We test whether \textsc{Muse}-derived probabilities improve model accuracy and calibration through supervised fine-tuning (SFT). We evaluate SFT variants that differ in how probabilities are injected into prompts and whether explicit reasoning is included. 

\paragraph{Direct SFT.}
We inject probabilistic signals via two prompt formats: (i) \textbf{Default}: include the \textsc{Muse} consensus \(\hat{p}\) with the patient input and gold label; (ii) \textbf{RawProb}: replace \(\hat{p}\) with bootstrapped per-model probabilities (e.g., Mistral: [0.62, 0.64, 0.60]; Qwen: [0.58, 0.55, 0.57]; DS-Qwen: [0.55, 0.53, 0.56]).

\paragraph{Chain-of-thought distillation.}
We optionally append teacher-generated reasoning using three variants: (i) \textbf{Original} (assess whether \(\hat{p}\) is reasonable; Table~\ref{tab:cot_prompts}), (ii) \textbf{Bayesian} (explicit Bayes framing), and (iii) \textbf{No \(\hat{p}\)} (reason without \(\hat{p}\)). Each CoT variant is paired with both SFT formats. 





\section{Experimental Setup}
We use \textit{TruthfulQA}\cite{lin2022truthfulqa}, a benchmark of adversarially designed questions with labeled truthful and untruthful answers. The task is to classify each candidate answer as truthful (\texttt{Yes}) or not (\texttt{No}), enabling direct evaluation of both discrimination (AUROC) and calibration (ECE, Brier Score). We also apply our method to clinical prediction tasks using two structured EHR datasets: diagnosis prediction from \textit{EHRShot}\cite{wornow2023ehrshot} and \textit{MIMIC-Extract}~\cite{wang2020mimic}. On MIMIC, the LLMs predict three outcomes: hospital length of stay $\geq$ 3 days (LOS3), $\geq7$ days (LOS7), and in-hospital mortality (Mort Hosp.). 

We evaluate the following open-source models: Mistral-7B-Instruct~\cite{jiang2023mistral7b}, Gemma-7B-it~\cite{team2023gemini}, Qwen2-7B-instruct~\cite{yang2024qwen2}, and the latest Deepseek-R1-Distil-Qwen-32B (DS-Qwen)~\cite{deepseekai2025deepseekr1incentivizingreasoningcapability}. All LLMs are run on a server with 4×A100 40GB GPUs. For DS-Qwen, we apply 8-bit quantization to reduce inference time and memory usage. 
In addition to single LLM method baseline, we compose two naive multi-LLM baselines: majority voting (counting positive labels), and mean of all LLMs' $\hat{p}^{yes}$ as the final positive probability.  
On the Uncertainty-aware SFT experiments, we investigate on two models, Mistral-7B-Instruct and Qwen2-7B-instruct, using the MIMIC-Extract dataset for the length-of-stay 3 days (LOS3) prediction task (\(y{=}1\) if LOS $\geq 3$ days, \(0\) otherwise). For CoT distillation, we leverage a HIPAA-Compliant Microsoft Azure GPT-o3-mini as the teacher model, and generate 300 samples on LOS-3 training set for each setting. This Azure GPT instances provides high-quality rationales while remaining compliant with the MIMIC-III data use agreement.
\begin{table}[t]
\small
\centering
\renewcommand{\arraystretch}{0.9}

\resizebox{\columnwidth}{!}{
\begin{tabular}{@{}lllllll@{}}
\toprule
\multicolumn{1}{c}{\textbf{MUSE}} & \multicolumn{1}{c}{\textbf{Task}} &
\multicolumn{1}{c}{\textbf{\begin{tabular}[c]{@{}c@{}}Total\\ \%\end{tabular}}} &
\multicolumn{1}{c}{\textbf{\begin{tabular}[c]{@{}c@{}}GEM\\ \%\end{tabular}}} &
\multicolumn{1}{c}{\textbf{\begin{tabular}[c]{@{}c@{}}Qwen\\ \%\end{tabular}}} &
\multicolumn{1}{c}{\textbf{\begin{tabular}[c]{@{}c@{}}DS-Qwen\\ \%\end{tabular}}} &
\multicolumn{1}{c}{\textbf{\begin{tabular}[c]{@{}c@{}}Mistral\\ \%\end{tabular}}} \\ \midrule
Conserv & LOS 3 & 31.6  & 56.97 & 35.67 & 25.01 & 17.42 \\
Greedy  & LOS 3 & 90.58 & 27.84 & 26.76 & 25.77 & 27.01 \\
Conserv & TQA   & 33.33 & 64.44 & 19.79 & 7.89  & 28.90 \\
Greedy  & TQA   & 76.18 & 33.46 & 26.89 & 17.17 & 31.95 \\ \bottomrule
\end{tabular}
}

\vspace{.5ex}

\begin{tabular}{@{}lllll@{}}
\toprule
\textbf{Method} & \textbf{Total\%} & \textbf{GEM\%} & \textbf{Qwen\%} & \textbf{DS-Qwen\%} \\ \midrule
Greedy   & 73.13 & 47.95 & 39.90 & 24.71 \\
Conserv. & 53.53 & 69.16 & 30.69 & 13.03 \\ \bottomrule
\end{tabular}

\vspace{-.1in}
\caption{\small Breakdown of LLM inclusion in \textsc{Muse}. Top: frequency of each model’s selection under Conservative and Greedy variants on LOS 3 and TruthfulQA (TQA). Bottom: inclusion frequencies when Mistral is removed from the model pool. (GEM: Gemma-7B, Qwen: Qwen2, DS-Qwen: Deepseek-distilled-Qwen32B, Mistral: Mistral-7B-Instruct).}
\label{tab:muse_inclusion_combined}
\end{table}


\begin{figure*}
    \centering
    \includegraphics[width=\textwidth]{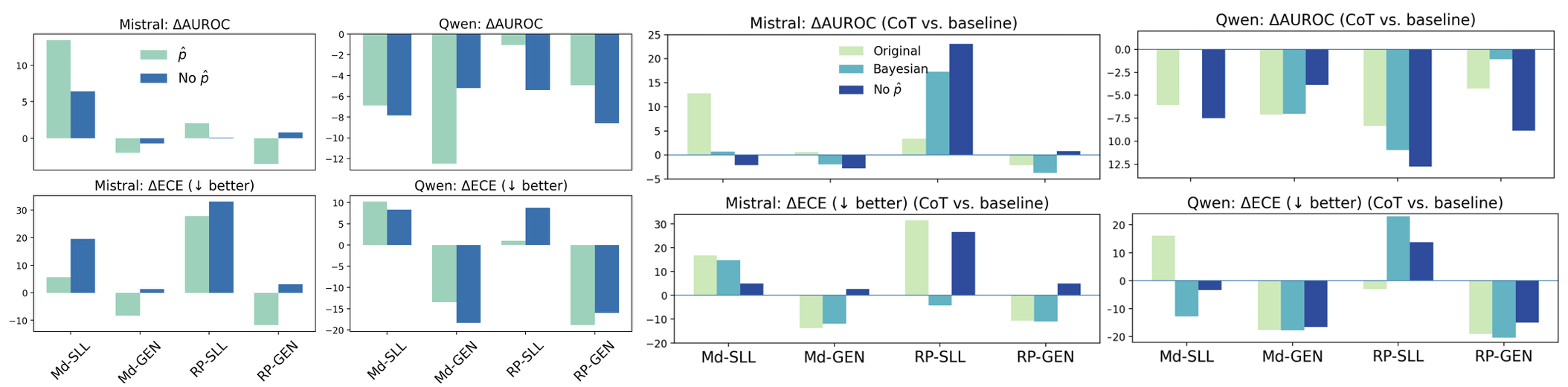}
    \vspace{-.26in}
    \caption{\small Comparative results of supervised fine-tuning with MUSE-derived probabilities. Common settings at both panels: Md indicates the default consensus probability input, while RP uses the raw bootstrapped probabilities from the model pool. Left (4 panels): Direct SFT performance shown as changes in AUROC and ECE (bottom row) for Mistral and Qwen, when using model SLL and GEN output under settings with and without $\hat{p}$. Improvements are measured relative to no-SFT baselines (positive $\Delta$AUROC, negative $\Delta$ECE indicate gains). Right (4 panels): CoT SFT performance under the same models, comparing three prompting strategies (Original, Bayesian, No $\hat{p}$). Results demonstrate that while direct SFT yields modest and model-dependent improvements, CoT-based SFT produces more variable outcomes across prompting strategies.}
    \label{fig:sft_cot_results}
\end{figure*}

\section{Results and Discussion}
We organize results around two questions.  
\textbf{Q1: Is \textsc{Muse} effective for uncertainty estimation?}  
We evaluate calibration and robustness on general and clinical datasets.  
\textbf{Q2: Can \textsc{Muse} provide silver-standard supervision for probabilistic reasoning?}  
We test whether consensus-derived probabilities aid SFT and CoT distillation.  
\noindent Findings show \textsc{Muse} improves calibration and robustness (Q1), while supervision results are mixed: some models benefit, others do not. Overall, multi-LLM aggregation is promising for uncertainty estimation but presents challenges to improve single-model reasoning.

\subsection{MUSE Effectiveness}
\textsc{Muse} improves both AUROC and calibration metrics compared to single LLMs and naive ensembling baselines, demonstrating its effectiveness in producing reliable and well-calibrated predictions through selective multi-model aggregation. Although the SLL method occasionally yields the highest AUROC, such as DS-Qwen achieving 72.89 on \textit{TruthfulQA}, it often suffers from poor calibration (ECE 57.30, Brier 54.16). In contrast, \textsc{Muse} offers more balanced predictions, with comparable AUROC (72.35) and substantially lower calibration error (ECE 38.15). 
Similar gains are observed on the EHRShot \textit{acute\_mi} task, where DS+Mistral+Qwen achieves the best Brier Score (24.4) and strong AUROC (62.2), improving over all single-LLM baselines.

\noindent \textbf{Demonstrating MUSE Impact.}
We address two-subquestions that illustrate the practical significance of the MUSE algorithm. 

\noindent \textbf{(1) Does \textsc{Muse} blindly include weaker models?}  
Table~\ref{tab:muse_inclusion_combined} reports selection frequencies under MUSE (Conservative, Greedy). Mistral is chosen less often than Greedy, only when it improves subset consistency. Even strong models (e.g., Qwen) are not always selected, indicating instance-specific, uncertainty-driven choice rather than global accuracy. When a weaker model adds noise, MUSE adapts. Removing Mistral shifts weight to GEM and Qwen, confirming adaptive selection rather than identity-based filtering.  

\noindent \textbf{(2) Are gains just from excluding a bad model?}  
Ablations over fixed subsets (Tables~\ref{tab:truthfulqa_llm_results}, \ref{tab:multi_llm_results}) and exhaustive pairwise/three-way combinations (Table~\ref{tab:adaptive_behavior_more}) show that while naive ensembles (e.g., DS+Qwen2) can perform well, MUSE often matches or exceeds them, even when weaker models remain, confirming that it performs input-level rather than fixed global selection. Additional analysis of per-model divergence from the consensus (Table~\ref{tab:jsd}) shows small JSD values ($<0.1$), supporting our claim that ensemble diversity with selective aggregation underlies MUSE’s effectiveness.



\subsection{MUSE-Guided Supervised Fine-Tuning}

Figure~\ref{fig:sft_cot_results} summarizes the impact of incorporating MUSE-derived probabilities into supervised fine-tuning.  Direct SFT yields modest, model-dependent effects. For Mistral, $\hat{p}$ sometimes improves discrimination but often worsens calibration, while GEN consistently lowers ECE. Qwen shows the opposite: calibration improves, but AUROC gains are limited or negative. Adding CoT distillation introduces wider variation. Mistral sees large AUROC increases with RawProb, though often at the cost of calibration. Qwen benefits most from Bayesian CoT, with clear calibration gains in GEN. The results highlight that \textsc{Muse}-derived uncertainty can provide useful supervision, but effects depend on model, supervision style, and whether signals are contextualized through reasoning.

\section{Conclusion}
We present \textsc{Muse}, a multi-LLM framework for uncertainty estimation that aggregates predictive distributions into calibrated, uncertainty-aware outputs. Beyond improving accuracy and calibration, we explored using \textsc{Muse} as a supervisory signal for fine-tuning single LLMs, with mixed but promising results, indicating a direction to explore further. 

\section*{Acknowledgments}

This work is supported by U.S. National Library of Medicine R00 LM014308.  

\section*{Limitation}
Our study evaluates a limited set of open-source LLMs and focuses exclusively on binary prediction tasks, where evaluation of discrimination and calibration is most straightforward. We also assume access to all model outputs during inference, which may not reflect real-time or resource-constrained deployment scenarios. However, our focus is not on maximizing efficiency, but on understanding how model composition and selective aggregation affect uncertainty estimation. Nonetheless, we have provided empirical evidence that \textsc{Muse} consistently improves both accuracy and calibration, highlighting the value of principled multi-model fusion. Future work will extend to more complex prediction settings and explore efficient selection strategies across broader model ecosystems. 

\section*{Ethical Consideration}
This study uses two publicly available, de-identified clinical datasets (MIMIC-Extract and EHRShot), ensuring no personally identifiable information is accessed or exposed. All models used are open-source LLMs, and no fine-tuning or data logging was performed, eliminating the risk of patient data leakage. While our focus is on evaluating uncertainty and not clinical deployment, we emphasize the need for responsible use of LLMs in sensitive domains. Any generative outputs or predictions from these models should be interpreted with caution, especially in clinical contexts, and subject to domain expert validation prior to real-world application.

\bibliography{custom}

\appendix

\section{More Analysis and Results.}

\subsection{\textsc{Muse} Conservative Algorithm}
\begin{algorithm}[t]
\footnotesize
\caption{\small \textsc{Muse}-Conservative version}
\label{alg:muse_conservative}
\begin{algorithmic}[1]
\REQUIRE Prediction set $\mathcal{P} = \{p_i\}_{i=1}^N$, confidence $c_i = |p_i^{\text{yes}} - 0.5|$, parameters $\beta$, $\tau$, $m_{\min}$
\STATE Sort $\mathcal{P}$ by $c_i$ descending; initialize $\mathcal{S} \gets \{p_1\}$, $u_{\text{total}}^{\text{prev}} \gets \infty$
\FOR{each $p_j$ in sorted $\mathcal{P} \setminus \mathcal{S}$}
  \STATE $\mathcal{S}' \gets \mathcal{S} \cup \{p_j\}$, $\bar{p} \gets \text{mean}(\mathcal{S}')$
  \STATE $u_{\text{epis}} \gets \frac{1}{|\mathcal{S}'|} \sum_{p \in \mathcal{S}'} \text{JS}(p \parallel \bar{p})^2$
  \STATE $u_{\text{alea}} \gets \frac{1}{|\mathcal{S}'|} \sum_{p \in \mathcal{S}'} H(p)$
  \STATE $u_{\text{total}} \gets u_{\text{epis}} + \beta \cdot u_{\text{alea}}$
  \IF{$|\mathcal{S}'| \geq m_{\min}$ and $u_{\text{total}} > u_{\text{total}}^{\text{prev}} - \tau$}
    \STATE \textbf{break}
  \ENDIF
  \STATE $\mathcal{S} \gets \mathcal{S}'$, $u_{\text{total}}^{\text{prev}} \gets u_{\text{total}}$
\ENDFOR
\STATE $\hat{p}^{\text{yes}} \gets \text{mean}_{p \in \mathcal{S}}(p^{\text{yes}})$
\RETURN $(\hat{p}^{\text{yes}}, u_{\text{total}}, \mathcal{S})$
\end{algorithmic}
\end{algorithm}

\begin{figure}[t!]
    \centering
    \includegraphics[width=\columnwidth]{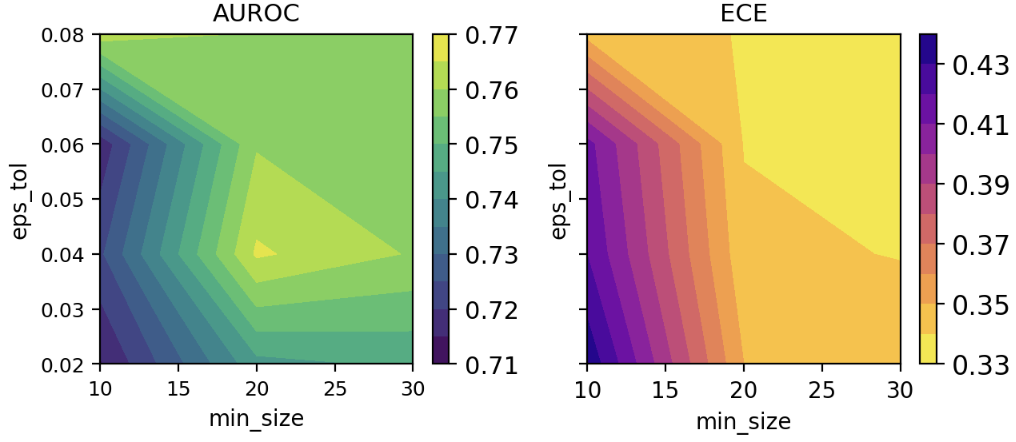}
    \vspace{-.28in}
    \caption{\small Contour plot of AUROC and ECE as \textsc{Muse} parameters ($m_{size}$, $\epsilon_{\text{tol}}$) vary, based on a TQA dev set. Main results use $m_{size}{=}20$, $\epsilon_{\text{tol}}{=}0.04$. See Appendix for further analysis. }
    \label{fig:param_sensitivity}
\end{figure}

\begin{table}[t!]
\centering
\small
\begin{tabular}{lccc}
\toprule
\textbf{Metric} & \textbf{P(Yes)} & $U(\mathcal{S})$  & \textbf{Best} \\
\midrule
\multicolumn{4}{l}{\textit{LOS 3}} \\
AUROC       & \textbf{0.5804} & 0.5103 & P(Yes) \\
ECE         & \textbf{0.2015} & 0.2208 & P(Yes) \\
Brier Score & 0.3321 & \textbf{0.3168} & $U(\mathcal{S})$ \\
\midrule
\multicolumn{4}{l}{\textit{LOS 7}} \\
AUROC       & \textbf{0.5748} & 0.5005 & P(Yes) \\
ECE         & \textbf{0.1998} & 0.2135 & P(Yes) \\
Brier Score & 0.3514 & \textbf{0.1519} & $U(\mathcal{S})$  \\
\midrule
\multicolumn{4}{l}{\textit{Mortality}} \\
AUROC       & \textbf{0.5598} & 0.5263 & P(Yes) \\
ECE         & \textbf{0.0283} & 0.1140 & P(Yes) \\
Brier Score & \textbf{0.1073} & 0.1085 & P(Yes) \\
\bottomrule
\end{tabular}
\caption{\small Comparison of predicted probability $p(\texttt{Yes})$ vs. total uncertainty (epistemic + aleatoric) as scoring signals. AUROC favors $p(\texttt{Yes})$, while Brier Score occasionally improves with uncertainty-based scoring.}
\label{tab:prob-vs-uncertainty}
\end{table}

\begin{table*}[t!]
\centering
\small
\renewcommand{\arraystretch}{0.6}
\begin{tabular}{llcccccccccc}
\toprule
\textbf{LLM} & \textbf{Method} & 
\multicolumn{3}{c}{\textbf{LOS3}} & 
\multicolumn{3}{c}{\textbf{LOS7}} & 
\multicolumn{3}{c}{\textbf{Mort Hosp.}} \\
\cmidrule(r){3-5} \cmidrule(r){6-8} \cmidrule(r){9-11}
& & \textbf{AUROC} & \textbf{ECE} & \textbf{Brier} 
  & \textbf{AUROC} & \textbf{ECE} & \textbf{Brier} 
  & \textbf{AUROC} & \textbf{ECE} & \textbf{Brier} \\
\midrule
Mistral-7B & SLL & 41.41 & 11.34 & 26.21 & 46.08 & 44.59 & 27.02 & 44.15 & 28.78 & 18.81 \\
           & GEN\textsuperscript{BS} & 51.28 & 15.63 & 28.37 & 55.49 & 38.96 & 25.38 & 52.70 & 14.59 & 14.07 \\
\midrule
Gemma-7B & SLL & 46.48 & 24.71 & 33.85 & 51.37 & 16.04 & 13.44 & 58.89 &	10.52& 10.48 \\
         & GEN\textsuperscript{BS} & 55.09 & 36.91 & 42.78 & 53.93 & 38.29 & 36.70 & 55.94	& 10.40 &15.45 \\ 
         \midrule 
\textsc{Muse} & Greedy  & \textbf{60.70} & 13.34 & 26.42 & 60.70 & 36.43 & 24.30 & 62.25 & 4.36 & 9.59 \\
             & weighted & 60.24 & 18.38 & 28.34 & \textbf{61.02} & 35.14 & 26.07 & \textbf{62.88} & 4.02 & \textbf{9.51} \\
             & Conserv.  & 58.34 & 30.87 & 36.17 & 58.74 & 36.29 & 32.15 & 61.89 & \textbf{3.88} & 10.90 \\
\bottomrule
\end{tabular}
\vspace{-.1in}
\caption{\small More AUROC, ECE, and Brier Score across clinical prediction tasks (MIMIC-Extract) for Mistral-7B and Gemma-7B. In this table, we also report the \textsc{Muse} results from Mistral, Gemma, DS-Qwen and Qwen.  }
\label{tab:mist_gem_mimic}
\end{table*} 

\begin{table}[t!]
\centering
\small
\begin{tabular}{lccc}
\toprule
\textbf{LLMs} & \textbf{AUROC} $\uparrow$ & \textbf{ECE} $\downarrow$ & \textbf{Brier Score} $\downarrow$ \\
\midrule
\multicolumn{4}{l}{\textit{Hyperlipidemia (weak models dominate)}} \\
Qwen & 46.91 & 38.92 & 50.88 \\
Mistral & 52.53 & 14.47 & 25.69 \\
Gemma & 45.18 & 25.72 & 35.36 \\
Deepseek-Distill & 43.92 & 22.46 & 40.97 \\
\midrule
Greedy (mean) & 33.16 & 24.51 & 29.13 \\
 (weighted) & 33.17 & 27.24 & 31.16 \\
\midrule
\midrule
\multicolumn{4}{l}{\textit{Lupus (encountered strong, stronger)}} \\
Qwen & 45.36 & 53.09 & 47.03 \\
Mistral & 33.89 & 19.38 & 11.39 \\
Gemma & 51.60 & 13.97 & 11.54 \\
Deepseek-Distill & 50.94 & 3.70 & 6.84 \\
\midrule
Greedy (mean) & 50.56 & 10.94 & 22.48 \\
 (weighted) & 52.45 & 11.82 & 23.09 \\
\bottomrule
\end{tabular}
\caption{Performance comparison across two EHRShot tasks. In \textit{hyperlipidemia}, where weak models dominate, the multi-LLM algorithm underperforms. In \textit{lupus}, encountering stronger base models allows the algorithm to adapt and perform competitively, reflecting the adaptive behavior pattern.}
\label{tab:contrast_adaptive_behavior}
\end{table}

\begin{table}[ht]
\centering
\small
\begin{tabular}{lcc}
\toprule
\textbf{LLMs} & \textbf{AUROC} & \textbf{AUROC} \\
Combination & Naive & \textsc{Muse}\\
\midrule
Qwen + Mistral & 56.78 & 55.17 \\
Mistral + Gemma & 55.14 & 55.15 \\
Gemma + Qwen & 51.87 & 54.20 \\
Mistral + Gemma + DS & 57.41 & 65.81 \\
Mistral + Gemma + Qwen & 54.60 & 56.47 \\
Qwen + Gemma + DS & 55.23 & 61.05 \\
\bottomrule
\end{tabular}
\caption{\small Performance on the EHRShot acute mi task across
different multi-LLM combinations. \textbf{AUROC Avg. Score} column is a simple
average of the bootstrapped score of individual LLMs.
\textbf{AUROC Algo. Score} signifies the (non-weighted) AUROC score of the combination of LLMs using Greedy approach.}
\label{tab:adaptive_behavior_more}
\end{table}

\begin{table}[]
\centering 
\small 
\begin{tabular}{@{}lll@{}}
\toprule
\textbf{Model on LOS3} & \textbf{JSD (Conserv)} & \textbf{JSD (Greedy)} \\ \midrule
GEM                    & 0.0444                 & 0.0709                \\
Qwen2                  & 0.0971                 & 0.0532                \\
DS-Qwen                & 0.0640                 & 0.06512               \\
Mistral                & 0.0948                 & 0.05126               \\ \bottomrule
\end{tabular}
\vspace{-.1in}
\caption{\small Comparison of JSD between individual LLM and final MUSE predicted probability. LLM notation is consistent with table~\ref{tab:muse_inclusion}}
\label{tab:jsd}
\end{table}

\begin{table}[ht!]
\centering
\begin{tcolorbox}[colback=gray!5, colframe=gray!40, width=0.95\linewidth]
\small
\textbf{Base CoT Prompt:} \\
Generate a short chain-of-thought (CoT) reasoning paragraph (maximum 200 words) 
that explains the significance of the given \texttt{p\_hat} value in the context of 
hospital length of stay (LOS) prediction. \\[0.5em]

\textbf{Bayesian CoT Prompt:} \\
Generate a short chain-of-thought (CoT) reasoning paragraph (maximum 200 words) 
that explains the significance of the given \texttt{p\_hat} value in the context of 
hospital length of stay (LOS) prediction. Use Bayesian reasoning to justify if 
\texttt{p\_hat} is a reasonable estimate. If \texttt{p\_hat} aligns with the true 
label, explain why it succeeds. If it does not align, explain why it fails. \\[0.5em]

\textbf{Additional Information (for both prompts):} \\
-- The \texttt{input} field contains clinical data from the MIMIC dataset. \\
-- The \texttt{p\_hat} value is the predicted probability of the patient staying 
three or more days. \\
-- The \texttt{y\_true} is the ground-truth label 
(0 = LOS $< 3$ days, 1 = LOS $\geq 3$ days). 
\end{tcolorbox}
\caption{\small Base and Bayesian chain-of-thought prompts used for supervised fine-tuning.}
\label{tab:cot_prompts}
\end{table}

\begin{figure*}[t!]
    \centering
    \includegraphics[width=\textwidth]{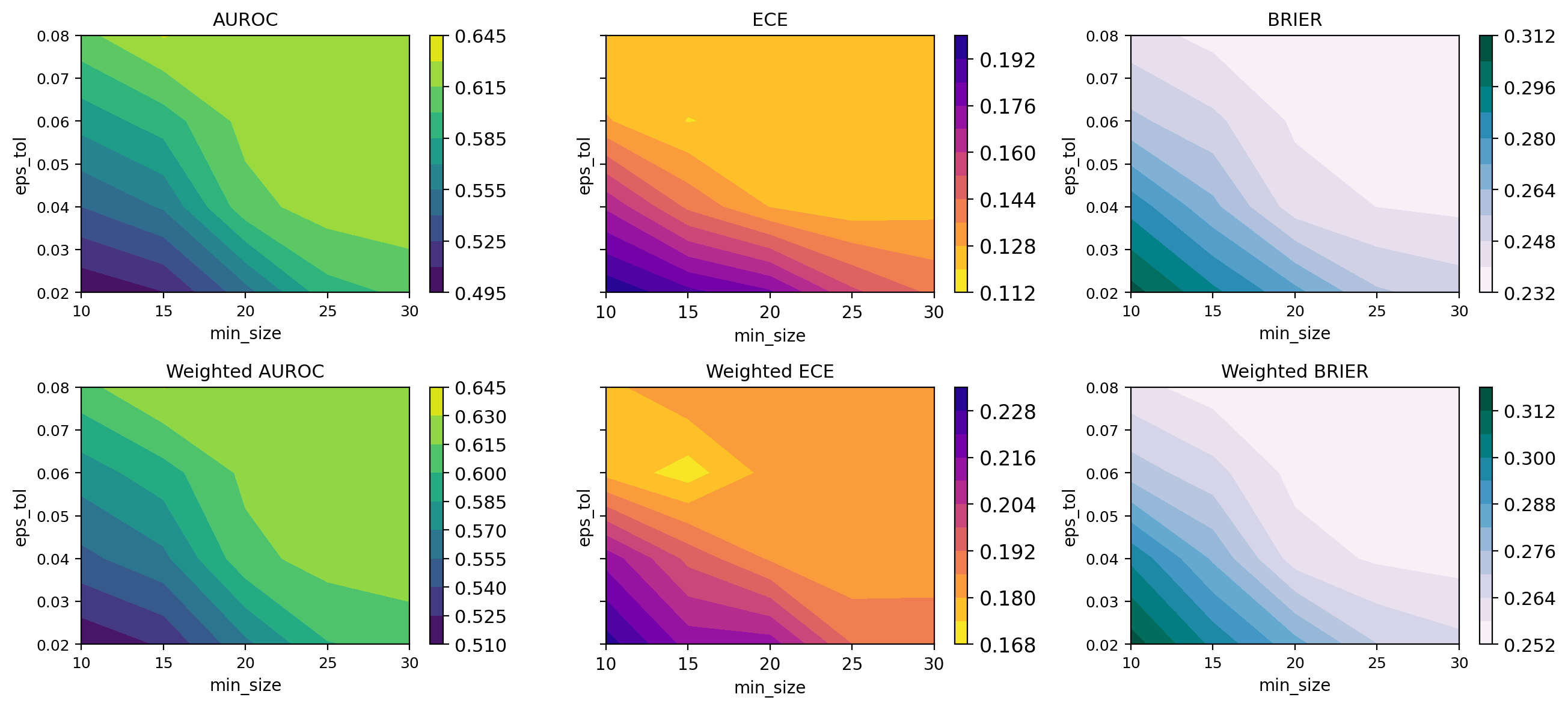}
    \caption{\small Contour plot for parameter sensitivity analysis using \textit{lupus} prediction task from EHRShot. We report \textsc{Muse}-Greedy with both weighted and unweighted version, to showcase the differences.}
    \label{fig:ehrshot_param}
\end{figure*}

\begin{table*}[ht!]
\centering
\small
\renewcommand{\arraystretch}{0.8}
\resizebox{\textwidth}{!}{
\begin{tabular}{@{}lll|ccc|ccc@{}}
\toprule
 & & & \multicolumn{3}{c|}{\textbf{With $\hat{p}$}} & \multicolumn{3}{c}{\textbf{No $\hat{p}$}} \\
\textbf{Model} & \textbf{SFT} & \textbf{Output} & AUROC$\uparrow$ & Brier$\downarrow$ & ECE$\downarrow$ & AUROC$\uparrow$ & Brier$\downarrow$ & ECE$\downarrow$ \\
\midrule
\multirow{4}{*}{Mistral}
 & \multirow{2}{*}{default} & SLL               & \textbf{54.83} & 39.51 & 31.84 & \textbf{47.82} & 61.17 & 45.69 \\
 &  & GEN\textsuperscript{BS} & 49.28 & 29.39 & \textbf{20.00} & \textbf{50.59} & 38.82 & 29.75 \\
 & \multirow{2}{*}{RP} & SLL                   & \textbf{43.46} & 53.75 & 54.04 & 41.48 & 59.28 & 59.28 \\
 &  & GEN\textsuperscript{BS} & 47.76 & 29.19 & \textbf{16.63} & \textbf{52.08} & 39.72 & 31.49 \\
\midrule
\multirow{4}{*}{Qwen}
 & \multirow{2}{*}{default}  & SLL               & 49.52 & 58.89 & 46.23 & 48.56 & 45.25 & 44.34 \\
 &  & GEN\textsuperscript{BS} & 44.31 & 49.97 & \textbf{20.70} & 51.58 & \textbf{37.43} & \textbf{15.82} \\
 & \multirow{2}{*}{RP}  & SLL & 55.35 & 38.94 & 36.99 & 51.00 & 46.54 & 44.77 \\
 &  & GEN\textsuperscript{BS} & 51.86 & 41.52 & \textbf{15.29} & 48.20 & 37.36 & \textbf{18.19} \\
\bottomrule
\end{tabular}
}
\vspace{.1in}
\caption{\small 
Direct SFT on Mistral-7B and Qwen2-7B under default and RawProb (RP) settings, with and without inclusion of $\hat{p}$. 
\textbf{Bold} indicates improvement over the no-SFT baselines: Mistral-7B: SLL 41.41/11.34/26.21, GEN\textsuperscript{BS} 51.28/15.63/28.37; Qwen2-7B: SLL 56.40/36.50/35.96, GEN\textsuperscript{BS} 56.80/31.10/34.18, reported in Tab~\ref{tab:llm_jsd_summary} and Tab~\ref{tab:mist_gem_mimic}. }
\label{tab:phat_results}
\end{table*} 

\begin{table*}[ht!]
\centering
\small
\setlength{\tabcolsep}{4pt}
\begin{tabular*}{\textwidth}{@{\extracolsep{\fill}} lll|ccc|ccc|ccc @{}}
\toprule
 & & & \multicolumn{3}{c|}{\textbf{Original}} & \multicolumn{3}{c|}{\textbf{Bayesian}} & \multicolumn{3}{c}{\textbf{No $\hat{p}$}} \\
\textbf{Model} & \textbf{SFT} & \textbf{Output} & {\small AUROC} & {\small Brier} & {\small ECE} & {\small AUROC} & {\small Brier} & {\small ECE} & {\small AUROC} & {\small Brier} & {\small ECE} \\
\midrule
\multirow{4}{*}{Mistral}
 & default  & SLL              & \textbf{54.22} & 42.90 & 42.97 & \textbf{42.14} & 43.04 & 41.03 & 39.27 & 35.09 & 31.23 \\
 &          & GEN\textsuperscript{BS} & \textbf{51.91} & 27.97 & \textbf{14.66} & 49.30 & 28.80 & \textbf{16.48} & 48.52 & 41.73 & 31.03 \\
 & RawProb  & SLL                  & \textbf{44.79} & 57.75 & 57.78 & \textbf{58.73} & 28.90 & \textbf{21.88} & \textbf{64.51} & 51.46 & 52.80 \\
 &          & GEN\textsuperscript{BS}     & 49.16 & 28.97 & \textbf{17.60} & 47.56 & 29.66 & \textbf{17.38} & 52.05 & 42.01 & 33.29 \\
\midrule
\multirow{4}{*}{Qwen}
 & default  & SLL              & 50.34 & 52.56 & 52.08 & 56.35 & \textbf{30.76} & \textbf{23.23} & 48.89 & 36.79 & 32.53 \\
 &          & GEN\textsuperscript{BS} & 49.72 & 44.55 & \textbf{16.63} & 49.77 & 43.65 & \textbf{16.49} & \textbf{52.95} & 37.29 & \textbf{17.59} \\
 & RawProb  & SLL                  & 48.07 & 36.16 & \textbf{32.97} & 45.43 & 58.95 & 59.01 & 43.63 & 51.21 & 49.74 \\
 &          & GEN\textsuperscript{BS}     & 52.55 & 40.98 & \textbf{15.19} & 55.72 & 39.83 & \textbf{13.81} & 47.94 & 40.25 & \textbf{19.23} \\
\bottomrule
\end{tabular*}
\vspace{.1in}
\caption{\small
SFT on Mistral-7B and Qwen2-7B with CoT prompts (Original, Bayesian, No $\hat{p}$), under Default/RawProb CoT and SLL/GEN\textsuperscript{BS}. 
\textbf{Bold} values indicate improvements over the no-SFT baselines (Qwen2-7B: SLL 56.40/36.50/35.96, GEN\textsuperscript{BS} 56.80/31.10/34.18; Mistral-7B: SLL 41.41/11.34/26.21; GEN\textsuperscript{BS} 51.28/15.63/28.37), reported in Tab~\ref{tab:llm_jsd_summary} and Tab~\ref{tab:mist_gem_mimic}. }
\label{tab:cot_results}
\end{table*}

Algorithm~\ref{alg:muse_conservative} presents the \textsc{Muse} conservative version. We consider total uncertainty as the sum of epistemic and aleatoric components to better balance diversity and reliability in model selection. The conservative version of \textsc{Muse} adopts a cautious strategy by only adding models when their inclusion leads to a meaningful reduction in total uncertainty. This prevents noisy or unstable predictions from being included, resulting in a more stable and selective ensemble that emphasizes trustworthy aggregation rather than maximizing diversity alone.

\subsection{Balancing diversity and noise.}
A key strength of our approach is its ability to balance diversity with reliability in multi-LLM ensembles. Figure~\ref{fig:param_sensitivity} shows that increasing the minimum subset size ($m_{\text{size}}$) and moderately relaxing the epistemic uncertainty threshold ($\epsilon_{\text{tol}}$) consistently improves both AUROC and ECE. A larger $m_{\text{size}}$ ($\geq$ 20) promotes diversity by including more models, while a moderate $\epsilon_{\text{tol}}$ ([0.04, 0.08]) allows controlled disagreement without overwhelming the ensemble with noise. The best performance is achieved when both parameters are carefully balanced. This supports our hypothesis that LLMs offer complementary strengths and provides empirical evidence for our subset-based uncertainty aggregation framework.  

\subsection{Adaptive model selection.}
Our method further demonstrates adaptive behavior: performance improves when “stronger” LLMs are present but degrades when weak or noisy models dominate the candidate pool, where strength is defined by each model’s single-task performance. This effect is most evident in \textit{acute\_{mi}} and \textit{TQA}. In EHRShot, DS and Mistral form a strong ensemble, but adding a weaker model like Gemma introduces noise that contaminates the pool and harms performance. This supports our hypothesis that selective aggregation for trustable consensus is the key to reliable ensemble performance. \textsc{Muse} makes no assumption that more models lead to better results; rather, it selectively aggregates those that contribute meaningful, calibrated signals to reduce total uncertainty. Table~\ref{tab:contrast_adaptive_behavior} includes a pair of contrasting cases from two prediction tasks in EHRShot for readers who's interested in (\textit{``weak models dominate''} vs. \textit{``encountered strong, stronger’’}) behavior.

\subsection{Comparison of P(Yes) vs. $U(\mathcal{S})$}
\label{app:total_uncertainty}

To compare different scoring strategies, we evaluate the predicted probability \( p(\texttt{Yes}) \) and the total uncertainty (the sum of epistemic and aleatoric components) as predictors of label correctness. As shown in Table~\ref{tab:prob-vs-uncertainty}, \( p(\texttt{Yes}) \) consistently achieves higher AUROC and lower ECE across all tasks, indicating better discrimination and calibration. However, total uncertainty yields lower Brier scores in some cases (e.g., LOS7), suggesting it may better reflect the overall confidence--error trade-off in noisier settings. These results indicate that while \( p(\texttt{Yes}) \) is a strong default for classification, total uncertainty can serve as a complementary signal for soft calibration or abstention.

\subsection{More results on MIMIC-Extract}

Table~\ref{tab:mist_gem_mimic} presents the performance of two single LLMs and the \textsc{Muse} multi-LLM approach across three clinical prediction tasks using three uncertainty estimation methods: sequence likelihood (SLL), generation-based prediction with bootstrapped generation (GEN\textsuperscript{BS}). The results show that \textsc{Muse}, particularly with the Greedy v2 strategy that minimizes total uncertainty, consistently improves AUROC while also reducing calibration error and Brier score compared to individual LLMs. For instance, in the Mortality prediction task, Greedy v2 achieves the highest AUROC (62.88) and the lowest Brier score (9.51), outperforming both Mistral and Gemma models under all methods. Similarly, in LOS3 and LOS7, \textsc{Muse} achieves competitive or best AUROC while offering substantial improvements in calibration, with ECE as low as 4.02 in Mortality. The Conservative variant further enhances calibration, reaching an ECE of 3.88, though at the cost of slightly lower AUROC. These findings demonstrate the effectiveness of \textsc{Muse} in producing more reliable and better-calibrated predictions by aggregating complementary strengths from multiple LLMs. 

\subsection{\textsc{Muse} JSD from the four LLMs} 
Table~\ref{tab:jsd} reports the average JSD between each LLM and the \textsc{Muse} consensus. Both weaker and stronger models show similarly small divergences, with no clear relation between a model’s standalone performance and its JSD to MUSE. 

\subsection{Analyzing the adapting behavior of \textsc{Muse} method via EHRShot}

Table~\ref{tab:contrast_adaptive_behavior} illustrates the adaptive behavior of our multi-LLM calibration algorithm. In the hyperlipidemia task, where all individual LLMs perform modestly, the aggregated model underperforms, indicating that combining weak predictors can degrade performance. In contrast, for the lupus task, where strong base models (e.g., Deepseek-Distill) are available, the algorithm adapts effectively, matching the best AUROC while maintaining good calibration. This contrast demonstrates the algorithm’s adaptive strength: it amplifies strong signals when present, but cannot compensate when no reliable model exists. 

Table~\ref{tab:contrast_adaptive_behavior} shows the impact of different LLM combinations on AUROC for the EHRShot acute MI task. While naive averaging yields modest performance gains, the \textsc{Muse} algorithm substantially boosts AUROC by selectively aggregating informative models. Notably, combinations with higher average AUROC do not always lead to better algorithmic performance: e.g., Qwen+Mistral ranks highest by average but is outperformed by combinations including DeepSeek when selected adaptively. This reinforces that performance is not simply a function of the number of models, but of their individual quality and how well their signals complement one another.

\subsection{Parameter sensitivity on EHRShot}

We evaluate how \textsc{Muse} performance varies with the two key hyperparameters: minimum subset size ($m_{\text{size}}$) and epistemic tolerance ($\epsilon_{\text{tol}}$), using EHRShot as the evaluation dataset (\textit{lupus} prediction). The top row shows unweighted AUROC, ECE, and Brier scores, while the bottom row shows the same metrics when aleatoric uncertainty is used as weighting in aggregation.

Overall, larger $m_{\text{size}}$ and moderate $\epsilon_{\text{tol}}$ (0.04–0.08) consistently lead to better performance across all metrics. The gains are especially pronounced in calibration (lower ECE and Brier), showing the benefit of including diverse yet coherent model outputs. Aleatoric weighting further improves stability, particularly under looser inclusion criteria. These trends confirm that careful tuning of subset size and disagreement tolerance is key to balancing diversity and reliability in multi-LLM ensembles.

\subsection{Prompting Strategy for Chain-of-Thought}
Table \ref{tab:cot_prompts} shows the prompts used for generating the base and Bayesian chain-of-thought reasoning paragraphs.

\subsection{Detailed results of MUSE-guided SFT}

When comparing against the no-SFT baselines, Direct SFT shows mixed improvements. For Mistral, SLL with $\hat{p}$ markedly improves AUROC (41.41 $\to$ 54.83), though calibration deteriorates (ECE 26.21 $\to$ 31.84). GEN\textsuperscript{BS} settings yield the strongest calibration benefits, lowering ECE to 16.63 compared to the baseline 28.37, while maintaining competitive AUROC (52.08). For Qwen, SLL with RawProb achieves AUROC (55.35) close to baseline (56.40) but with worse calibration (ECE 36.99 vs.\ 34.18). By contrast, GEN\textsuperscript{BS} achieves notable calibration gains, reducing ECE to 15.29--18.19 compared to the baseline 34.18, though AUROC remains lower. Overall, Direct SFT demonstrates that introducing $\hat{p}$ or raw probability signals can help calibration, especially under GEN\textsuperscript{BS}, but often at the cost of discrimination.  

\paragraph{CoT distillation SFT (Table~\ref{tab:cot_results}).}  
Adding teacher-generated reasoning further diversifies the outcomes. For Mistral, Bayesian RawProb SLL produces the highest AUROC (58.73 vs.\ baseline 41.41), while No~$\hat{p}$ RawProb reaches 64.51, albeit with very poor calibration (ECE 52.80). GEN\textsuperscript{BS} again proves more stable, with Original CoT reducing ECE to 14.66, and Bayesian CoT to 16.48, compared to the baseline 28.37. For Qwen, Bayesian Default SLL achieves balanced improvement, raising AUROC to 56.35 while reducing ECE to 23.23 (baseline 34.18). Similarly, GEN\textsuperscript{BS} with RawProb Bayesian CoT yields the lowest calibration error overall (ECE 13.81). These findings suggest that CoT distillation can help models internalize probabilistic reasoning and achieve strong calibration improvements, though sometimes at the expense of discrimination (e.g., Qwen SLL RawProb). 

The results highlight that \textsc{Muse}-derived uncertainty can provide useful supervision, but the benefits are strongly dependent on LLM, supervision style, and whether the probabilistic signals are contextualized through reasoning.

\end{document}